\documentclass[11pt]{article}

\usepackage[final]{acl}

\usepackage{times}

\usepackage{fontspec}

\newfontfamily{\gurmukhi}[
  Path=fonts/,
  Extension=.ttf,
  UprightFont=NotoSansGurmukhi-Regular
]{NotoSansGurmukhi}

\newfontfamily{\shahmukhi}[
  Path=fonts/,
  Extension=.ttf,
  UprightFont=NotoSansArabic-Regular
]{NotoSansArabic}

\usepackage[T1]{fontenc}
\usepackage{enumitem}
\usepackage{xcolor}
\definecolor{blue01}{RGB}{0, 0, 150}
\usepackage{microtype}

\usepackage{inconsolata}
\usepackage{booktabs} 
\usepackage{tcolorbox} 
\usepackage{graphicx}
\usepackage{tabularx}
\usepackage{amsmath}   
\usepackage{amssymb}   
\usepackage{latexsym}
\usepackage{tcolorbox}
\usepackage{booktabs}
\usepackage{tabularx}
\usepackage{colortbl}
\usepackage{xcolor}
\usepackage{ragged2e}
\usepackage{array}
\usepackage{graphicx}
\usepackage[table]{xcolor}
\usepackage{microtype}
\usepackage{hyperref}
\usepackage[capitalize]{cleveref}
\usepackage{multirow}
\usepackage{float}

\crefname{appendix}{App.}{App.}
\definecolor{best}{HTML}{0000FF} 
\definecolor{worst}{HTML}{FF0000} 
\definecolor{headergray}{HTML}{F2F2F2}
\definecolor{rowgray}{HTML}{F9F9F9}
\definecolor{gray01}{gray}{0.3}
\title{Not Truly Multilingual: Script Consistency as a Missing Dimension in VLM Evaluation\thanks{Data and code is available at \url{https://github.com/prabhjotschugh/Not-Truly-Multilingual-PuMVR}.}}

\author{Prabhjot Singh\textsuperscript{$\dagger$, $\S$} \quad
    Bhushan Pawar\textsuperscript{$\dagger$} \quad
    Madhu Reddiboina\textsuperscript{$\dagger$} \quad
    Rajvee Sheth\textsuperscript{$\ddagger$} \\ 
  \\
  \textsuperscript{$\dagger$}RediMinds Inc., USA \\
  \textsuperscript{$\S$}The University of Texas at Austin, USA \\
  \textsuperscript{$\ddagger$}Indian AI Research Organization (IAIRO), India \\
  \\
  \href{mailto:prabhjot.singh@utexas.edu}{prabhjot.singh@utexas.edu}
}

\begin{document}
\maketitle
\begin{abstract}
Current multilingual evaluations for Vision-Language Models (VLMs) assume a one-to-one mapping between language and orthography, overlooking billions of users of multi-script languages. We introduce \textbf{PuMVR} (Punjabi Multimodal Visual Reasoning), a benchmark of 1,000 strictly parallel image-text instances across Punjabi's three active scripts: Gurmukhi, Shahmukhi, and Roman. Evaluating 10 state-of-the-art VLMs, we expose a substantial and systematic \textbf{Script Gap}. Models frequently solve visual tasks in one script while failing identical tasks in another, with accuracy deltas reaching 16\%. Crucially, visual input boosts absolute performance uniformly yet does not close the orthographic gap. Furthermore, cross-script in-context transfer is highly brittle, exposing script-locked knowledge representation. Supported by McNemar tests across all script pairs, our findings demonstrate that current "multilingual" VLMs are not truly multi-script. We propose the \textbf{Script Consistency Rate (SCR)}, which falls as low as 24.8\% on our benchmark, as a mandatory metric for script-agnostic evaluation to ensure equitable AI access.
\end{abstract}

\section{Introduction}

The rapid evolution of Multimodal Vision-Language Models (VLMs) has been accompanied by claims of broad linguistic competence. Frontier systems such as GPT-4o \citep{openai2024gpt4o} and Gemini 1.5 \citep{reid2024gemini} are validated on their ability to reason across dozens of languages. However, these evaluations rest on a precarious assumption: the \textbf{``One Language, One Script'' (OLOS) paradigm}. By treating orthography as a deterministic function of language, current benchmarks, including XM3600 \citep{thapliyal2022crossmodal} and MaXM \citep{zhao2024maxm}, overlook the reality of hundreds of millions of users who navigate the world through multi-script systems.

This evaluation gap reveals important questions about model explainability. If a model's reasoning fluctuates when identical semantic content is presented in different scripts, its knowledge may rely more heavily on orthographic patterns than language-level semantic concepts. Script variation serves as a critical diagnostic probe: if a visual concept understood in one script is inaccessible in another, the model's ``multilinguality'' may rely more on pattern matching than purely semantic grounding.

This work exposes an equity gap: benchmarks assume script-language isomorphism, which misrepresents performance for script-switching users. For a Punjabi speaker using Roman and Gurmukhi, a model that succeeds in one script but fails in the other is not partially capable, it is unreliable. We introduce Script Consistency Rate (SCR) to reframe multilingual evaluation from language breadth to orthographic robustness, establishing script-agnosticism as a requirement for true multilingual AI.

We use Punjabi as a diagnostic case study because it uniquely satisfies three conditions for isolating script as an independent variable. Spoken by over 125 million people, Punjabi operates through three distinct systems: Gurmukhi (Indic script), Shahmukhi (Perso-Arabic), and Roman (Latin transliteration). We introduce \textbf{PuMVR} (Punjabi Multimodal Visual Reasoning), a parallel-script benchmark of 1000 expert-curated tasks isolating orthography as an independent variable in multimodal reasoning. As a single-language study, our empirical claims are scoped to Punjabi; we treat Punjabi as a diagnostic case for a broader class of multi-script languages rather than as direct evidence about those other languages. \\


\noindent\textbf{Our contributions are:} 
\begin{enumerate}

\item \textbf{The PuMVR Benchmark}: 1000 parallel instances across three scripts, providing the first controlled, three-way orthographic evaluation for multimodal reasoning. 
\item \textbf{Quantification of Script Bias}: A systematic audit of 10 state-of-the-art VLMs revealing significant performance gaps (accuracy deltas up to 16\%, Script Consistency Rates as low as 24.8\%) and limited cross-script transferability. 
\item \textbf{Statistical Validation of Script Bias}: McNemar tests across all pairwise script comparisons confirm the Script Gap is statistically robust - Gurmukhi-Shahmukhi gaps reach significance for 8 of 10 models, with 5 models at $p<0.001$.
\end{enumerate}

\section{Related Work}

The rise of VLMs, from CLIP \citep{radford2021learning} to instruction-tuned systems such as LLaVA \citep{liu2024llavanext} and Qwen-VL \citep{bai2023qwen}, has been accompanied by claims of multilingual competence across multiple languages. However, existing evaluation paradigms largely follow a 
one-script-per-language assumption. Current benchmarks conflate language with orthography, treating script as deterministic rather than variable. For over one billion speakers of digraphic languages like Punjabi ({\gurmukhi ਗੁਰਮੁਖੀ} (Gurmukhi)), {\shahmukhi شاہمکھی} (Shahmukhi), Roman), Serbian (Cyrillic, Latin), and Kurdish (Arabic, Latin, Cyrillic), this assumption masks orthographic bias that fragments AI access.

\subsection{Orthographic Gaps in Multilingual Multimodal Benchmarks}

Recent multilingual multimodal benchmarks have expanded evaluation coverage substantially. IGLUE \citep{bugliarello2022iglue}, XM3600 \citep{thapliyal2022crossmodal}, PaLo \citep{ma2024palo}, and MVL-SIB \citep{schmidt2025mvlsib} broadened cross-lingual evaluation across dozens to hundreds of languages, while MaRVL \citep{liu2021visually}, BLEnD-Vis \citep{zhang2025blendvis}, IndicVisionBench \citep{singh2025indicvisionbench}, and ALM-Bench \citep{vayani2025all} introduced culturally grounded multilingual reasoning tasks.

However, these milestones share a fundamental limitation: \textit{each language appears in exactly one script}. MaRVL represents Tamil in Tamil script and Swahili in Latin, but never tests whether Tamil speakers using Romanization experience equivalent performance. This OLOS assumption artificially inflates multilingual capabilities, a model achieving 85\% on "Punjabi" in Gurmukhi may drop to 69\% in Shahmukhi for identical content, a disparity invisible on current leaderboards. 


\subsection{The Script Gap: From Text-Only Evidence to Multimodal Urgency}

Text-only NLP has documented substantial script-dependent degradation. \citet{pfeiffer2021unks} showed that multilingual BERT fails on unseen scripts, while \citet{rust2021good} demonstrated inequitable tokenizer allocation across orthographies, with low-resource scripts often receiving significantly fewer subword units. These disparities have practical consequences: \citet{gupta2025script} reported 5--12 point F1 degradation for Romanized Hindi-Urdu healthcare queries. Although prior work has explored romanization effects \citep{amrhein2020romanization} and cross-script training strategies \citep{ng2024cori}, no existing work systematically evaluates whether multimodal grounding mitigates or amplifies script-dependent failures.

\subsection{Cultural Grounding and Script-Locked Knowledge}
Effective multimodal reasoning requires cultural grounding beyond object recognition \citep{yin2021broaden}. Prior work has shown that VLMs struggle with culturally specific concepts and reasoning \citep{liu2021visually, zhang2025blendvis}. However, existing benchmarks do not examine whether cultural knowledge itself becomes \textit{script-dependent}. This question is particularly important for multiscript languages such as Punjabi, where Gurmukhi, Shahmukhi, and Roman scripts are associated with distinct historical, religious, and sociocultural contexts. PuMVR enables systematic evaluation of whether VLMs activate different knowledge representations based solely on script variation, though isolating this from pretraining-corpus confounds requires care.



PuMVR addresses this gap by isolating script as an independent 
variable, introducing SCR and Transfer Efficiency (TE) as metrics 
for orthographic robustness, and providing a replicable methodology 
for multi-script evaluation.

\begin{figure}[H]
\centering
\begin{tcolorbox}[
    colback=blue!5!white, 
    colframe=blue!75!black, 
    title=Example ID: I\_1 , 
    fonttitle=\bfseries,
    left=6pt, right=6pt, top=4pt, bottom=4pt,
    boxsep=2pt,
    sharp corners=south
]
\small

\centering
\setlength{\fboxsep}{0pt} 
\setlength{\fboxrule}{0.5pt} 
\fbox{\includegraphics[width=0.45\columnwidth]{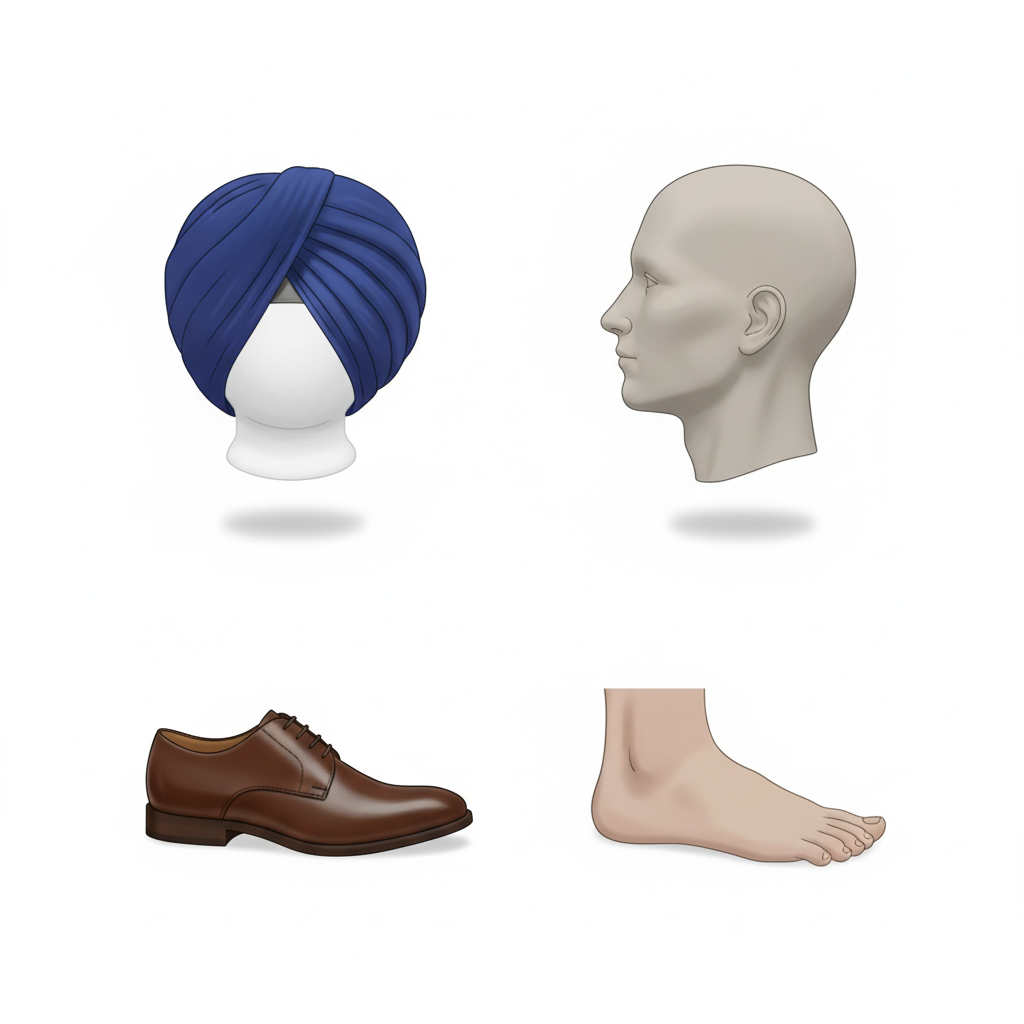}} 
\vspace{0.2cm}

\flushleft
\textbf{Gurmukhi:} {\gurmukhi ਪੱਗ : ਸਿਰ :: ਜੁੱਤੀ : ?} \\
\vspace{0.1cm}
A {\gurmukhi ਹੱਥ} \quad B {\gurmukhi ਪੈਰ} \quad C {\gurmukhi ਅੱਖ} \quad D {\gurmukhi ਨੱਕ} \\[10pt] 

\textbf{Shahmukhi:} {\shahmukhi پگ : سر :: جُتی : ؟} \\
\vspace{0.1cm}
A {\shahmukhi ہتھ} \quad B {\shahmukhi پیر} \quad C {\shahmukhi اکھ} \quad D {\shahmukhi ناک} \\[10pt] 

\textbf{Roman:} Pagg : Sir :: Jutti : ? \\
\vspace{0.1cm}
\textbf{A} hath \quad \textbf{B} pair \quad \textbf{C} akkh \quad \textbf{D} nakk \\[10pt] 

\textbf{Correct Answer:} \textbf{B} \\
\vspace{0.2cm}
\textbf{Reasoning:} Since a turban (\textit{pagg}) is worn on the head (\textit{sir}), just as a shoe (\textit{jutti}) is worn on the foot (\textit{pair}).
\end{tcolorbox}
\caption{One illustrative instance from PuMVR (not representative of the full item distribution) showing cross-script equivalent options.}
\label{fig:example}
\end{figure}

\section{The PuMVR Benchmark}

We introduce \textbf{PuMVR} (Punjabi Multimodal Visual Reasoning), the first benchmark designed to isolate script as an independent variable in multimodal reasoning. Unlike traditional multilingual benchmarks that conflate language with orthography, PuMVR provides 1000 culturally grounded image-reasoning tasks, each existing in perfect semantic equivalence across {\gurmukhi ਗੁਰਮੁਖੀ} (Gurmukhi), {\shahmukhi شاہمکھی} (Shahmukhi), and Roman scripts.

\subsection{Design Rationale: Why Punjabi?}

Punjabi serves as an ideal diagnostic for script bias due to three critical properties. 
\begin{itemize}[noitemsep]
    \item Its three active scripts, Gurmukhi (Indian Punjab, ~50M speakers), Shahmukhi (Pakistani Punjab, ~60M speakers), and Roman (diaspora/digital, ~15M users), are geographically and culturally isolated, minimizing training data overlap. 
    \item The scripts represent fundamentally different visual systems: Gurmukhi is an Indic abugida, Shahmukhi uses Perso-Arabic cursive, and Roman relies on Latin linearity.
    \item Punjabi remains low-resource compared to Hindi or Urdu, amplifying reliance on memorized patterns over semantic grounding.
\end{itemize}
Crucially, certain cultural concepts carry script-specific associations, Gurmukhi predominates in Sikh contexts (Golden Temple), while Shahmukhi aligns with Islamic heritage (Badshahi Mosque), enabling tests of script-locked knowledge retrieval.

\subsection{Dataset Composition and Structure}

PuMVR comprises 1,000 parallel instances. Each instance consists of one image paired with a question and four multiple-choice options 
provided in Gurmukhi, Shahmukhi, and Roman scripts, along with the corresponding correct answer (semantically equivalent across scripts); 
Figure~\ref{fig:example} shows a single illustrative instance and is not representative of the full item distribution. PuMVR spans multiple 
question types (visual analogies, cultural object recognition, festival and celebration reasoning, architectural and landmark recognition, 
text-in-image reasoning, and abstract visual-linguistic reasoning), several of which require the image to answer correctly and cannot be 
solved from the question text alone; the analogy item shown in Figure~\ref{fig:example} is one such format but is not solvable from world 
knowledge for every instance in the benchmark.

Each instance contains: an image, question text in all three scripts, four multiple-choice options per script, the correct answer per script, and a human-authored reasoning explanation. By making script an explicit, controllable variable, the methodology extends naturally to other multi-script languages including Hindi-Urdu, Serbian, Kurdish, and Sindhi.

This structure preserves the same visual and semantic content across scripts, allowing observed differences in model behavior to be attributed primarily to orthographic variation rather than task difficulty or dataset composition.


\subsection{Annotation Efforts and Quality}

To validate the quality and reliability of \textsc{PuMVR}, we conducted a formal annotation quality study. The pipeline involved four annotators: two dataset curators (one of whom is an author of this paper), who were responsible for initial instance creation, and two independent, paid quality-check annotators recruited specifically for verification. All four participants are native speakers of Punjabi or Urdu and are fluent in the scripts they evaluated.

\paragraph{Annotation Guidelines.}
Comprehensive guidelines were prepared before any curation and annotation process, covering two complementary roles: (1)~\emph{curation guidelines} for instance authoring and semantic equivalence standards; and (2)~\emph{quality-checking guidelines} for evaluating instances and script correctness without reference to the original author's judgments. This dual approach ensured a shared standard and high-quality annotations.

\paragraph{Annotator Profiles.}
Both quality-check annotators were native Punjabi or Urdu speakers from Pakistan, compensated appropriately, and had no prior access to the benchmark instances or hypotheses. One annotator was fluent in Shahmukhi and Roman Punjabi/Urdu; the other was fluent in all three scripts (Gurmukhi, Shahmukhi, and Roman).

\paragraph{Annotation Protocol.}
All 1{,}000 instances were independently evaluated across five dimensions: semantic equivalence, answer correctness, and script accuracy for each of the three orthographies. Annotators applied a conservative criterion, marking \emph{No} under genuine uncertainty rather than defaulting to \emph{Yes}, ensuring that near-universal agreement reflects genuine expert consensus rather than passivity (see Appendix~\ref{app:annotation_protocol} for full guidelines).

\paragraph{Inter-Annotator Agreement.}
Inter-annotator agreement was consistently high across all five evaluation dimensions (Table~\ref{tab:iaa}). The label distribution is strongly skewed toward the positive class across all dimensions, reflecting the dataset’s curation process rather than annotator behavior. In such settings, chance-corrected agreement coefficients can become less informative due to prevalence effects. We therefore report Prevalence-Adjusted Bias-Adjusted Kappa (PABAK) \citep{byrt1993bias}, along with positive-class F1 and observed agreement ($P_o$), as the primary agreement metrics.


\begin{table}[t]
\centering
\small
\setlength{\tabcolsep}{3pt}
\renewcommand{\arraystretch}{1.2}
\begin{tabularx}{\columnwidth}{>{\RaggedRight\arraybackslash}X c c c}
\toprule
\textbf{Dimension} & \textbf{$P_o$} & \textbf{PABAK} & \textbf{F1} \\
\midrule
Semantic Equivalence & 99.8\% & 0.996 & 0.999 \\
Answer Correctness   & 98.5\% & 0.970 & 0.992 \\
Script Acc. - Gurmukhi  & 100\% & 1.000 & 1.000 \\
Script Acc. - Shahmukhi & 100\% & 1.000 & 1.000 \\
Script Acc. - Roman     & 100\% & 1.000 & 1.000 \\
\bottomrule
\end{tabularx}
\caption{Inter-annotator agreement across all 1,000 instances. We report observed agreement ($P_o$), Prevalence-Adjusted Bias-Adjusted Kappa (PABAK), and positive-class F1.}
\label{tab:iaa}
\end{table}

PABAK scores reach 0.970 or above and F1 scores reach 0.992 or above across all five dimensions. Script accuracy reached perfect agreement (100\%) across all three orthographies, confirming that orthographic correctness is unambiguous among native-script-proficient annotators. That these same instances expose accuracy deltas of up to 16\% and SCR values as low as 24.8\% across 10 VLMs validates PuMVR as a genuinely challenging benchmark - the annotators confirmed dataset quality, not task simplicity.

\section{Experimental Setup}

\begin{table}[ht]
\centering
\renewcommand{\arraystretch}{1.0}
\setlength{\tabcolsep}{6.5pt}
\begin{tabular}{@{}p{4.2cm} l@{}}
\toprule
\textbf{Model} & \textbf{Organization} \\
\midrule
\multicolumn{2}{@{}l@{}}{\textcolor{blue01}{\textbf{Proprietary Frontier Models}}} \\
gpt-4o & OpenAI \\
gemini-2.5-flash & Google DeepMind \\
claude-sonnet-4 & Anthropic \\
grok-4-1-fast-reasoning & xAI \\
\midrule
\multicolumn{2}{@{}l@{}}{\textcolor{blue01}{\textbf{Open-Weights Models}}} \\
Qwen2-VL-7B-Instruct & Alibaba \\
Qwen2-VL-72B-Instruct & Alibaba \\
Llama-3.2-11B-Vision & Meta \\
LLaVA-OneVision-1.5-8B-Instruct & LMMS-Lab \\
InternVL2\_5-26B & OpenGVLab \\
Kimi-VL-A3B-Instruct & Moonshot AI \\
\bottomrule
\end{tabular}
\caption{Evaluated VLMs spanning frontier and open-weights systems.}
\label{tab:models}
\end{table}

Our experimental framework examines the \textbf{Script-Reality Gap}  through three investigations in sequence: Experiment~1 establishes that script-dependent gaps exist and quantifies their magnitude; Experiment~2 tests whether visual input closes those gaps; and 
Experiment~3 tests whether in-context examples transfer knowledge across orthographies.

We evaluate 10 state-of-the-art VLMs spanning both proprietary and open-weight systems (Table~\ref{tab:models}), enabling assessment of frontier models alongside reproducible baselines. PuMVR contains 1,000 parallel instances; a stratified 375-instance evaluation split was fixed prior to any model evaluation to prevent data snooping, with the remaining 625 instances reserved for future fine-tuning and mitigation studies. Where a model produced an empty response for an instance (Appendix~\ref{app:error_analysis}), that instance is excluded from the denominator for the corresponding per-script accuracy cell; this affects a small number of cells, primarily for claude-sonnet-4 and Qwen2-VL-72B-Instruct.

\subsection{Experiment 1: The ``Script Gap'' Quantification}

\textbf{Objective}: Establish the existence and magnitude of script-dependent performance bias under identical semantic conditions.

\paragraph{Design}: Each PuMVR instance is evaluated in three isolated passes, one per script, to prevent cross-script priming. Models receive the image, question, and four options in a single script using script-specific instruction templates (\cref{appendix:prompts}), and must output the exact text of the correct option. We evaluate model performance using the following metrics.

\begin{enumerate}
    \item \textbf{Script Accuracy} ($\text{Acc}_s$): Raw per-script performance
    \begin{equation}
    \text{Acc}_s = \frac{1}{|I|} \sum_{i \in I} \mathbb{1}[\text{Pred}_s(i) = \text{GT}_s(i)]
    \end{equation}

    \item \textbf{Script Consistency Rate (SCR)}: Percentage of instances answered correctly across all three scripts simultaneously

    \begin{equation}
    \text{SCR} = \frac{1}{|I|} \sum_{i \in I} \prod_{s \in \mathcal{S}} \mathbb{1}[\text{Correct}_s(i)]
    \label{eq:scr_compact}
    \end{equation}
    where $\mathcal{S} = \{\text{Gur}, \text{Shah}, \text{Rom}\}$
    \\
    SCR serves as a strict script-agnostic benchmark. A model achieving 90\% per-script accuracy but only 70\% SCR reveals that 20\% of its knowledge is orthographically fragmented.

    \item \textbf{Performance Delta} ($\Delta$): Maximum accuracy variance
    \begin{equation}
    \Delta = \max_{s_1, s_2 \in S} |\text{Acc}_{s_1} - \text{Acc}_{s_2}|
    \end{equation}
\end{enumerate}

To confirm that this design measures comprehension rather than output formatting difficulty, we computationally verified all 11,250 model responses (10 models $\times$ 375 instances $\times$ 3 scripts). Across all scripts, 99.59\% of errors were complete wrong-option selections indicating semantic comprehension failures, 0.41\% were empty responses, and 0.00\% were formatting artifacts. These results confirm that observed performance gaps reflect orthographic comprehension failures rather than formatting limitations. Unlike per-script accuracy, SCR exposes orthographic fragmentation that leaderboards mask: a model with 90\% Gurmukhi and 85\% Shahmukhi accuracy may still serve fewer than 78\% of instances reliably across both scripts.


\subsection{Experiment 2: Modality Importance Ablation}

\textbf{Objective}: Determine whether visual grounding compensates for weak script comprehension or merely provides additive benefit.

\paragraph{Design}: We compare Text-Only (question + options, no image) versus Multimodal (complete input) conditions across all scripts.

\paragraph{Metric}: Visual Gain (VG) quantifies visual contribution:
\begin{equation}
\text{VG}_s = \text{Acc}_{\text{Multimodal}}(s) - \text{Acc}_{\text{Text-Only}}(s)
\end{equation}

Uniform VG across scripts indicates visual information does not close the script gap, the bias is systematic, not compensatory.

\subsection{Experiment 3: Cross-Script Transfer with Few-Shot Learning}
\label{sec:exp3}

\textbf{Objective}: Test whether in-context knowledge transfers across orthographies or remains script-locked.

\paragraph{Design}: We employ $k=3$ in-context exemplars under three conditions:

\begin{enumerate}
    \item \textbf{Monoscript}: Examples and test in same script (e.g., Gurmukhi $\to$ Gurmukhi)
    \item \textbf{Cross-Script}: Examples in different script (e.g., Roman $\to$ Gurmukhi)
    \item \textbf{Mixed-Script}: Examples rotated through all three scripts
\end{enumerate}

\textbf{Metrics}:

\begin{enumerate}
    \item \textbf{Few-Shot Lift (FSL)}: Improvement from zero-shot
    \begin{equation}
    \text{FSL}_{s} = \text{Acc}_{\text{fs}}(s \to s) - \text{Acc}_{\text{zs}}(s)
    \label{eq:fsl}
    \end{equation}
    where $\text{Acc}_{\text{fs}}$ and $\text{Acc}_{\text{zs}}$ denote the accuracies for few-shot and zero-shot conditions, respectively.

    \item \textbf{Transfer Efficiency (TE)}: Cross-script to in-script ratio
    \begin{equation}
    \text{TE}_{T \to S} = \frac{\text{Acc}_{\text{Few-Shot}}(T \to S)}{\text{Acc}_{\text{Few-Shot}}(S \to S)} \times 100\%
    \end{equation}
    
    TE quantifies how much in-context knowledge transfers across scripts. If TE $< 50\%$, the model has encoded script-specific surface patterns rather than transferable semantic representations. If TE exhibits asymmetry (e.g., $\text{TE}_{\text{Roman} \to \text{Gurmukhi}} \gg \text{TE}_{\text{Gurmukhi} \to \text{Roman}}$), it reveals an anchor script bias.
\end{enumerate}

\begin{table*}[t]
\centering
\small
\setlength{\tabcolsep}{2.5pt}
\renewcommand{\arraystretch}{1.15}
\begin{tabular}{@{}p{3 cm}l *{9}{l}@{}}
\toprule
\textbf{Model} & \multicolumn{3}{c}{\textbf{Monoscript}} & \multicolumn{3}{c}{\textbf{Cross-Script}} & \multicolumn{3}{c}{\textbf{Mixed-Script}} \\
\cmidrule(lr){2-4} \cmidrule(lr){5-7} \cmidrule(lr){8-10}
& \textbf{G$\to$G} & \textbf{S$\to$S} & \textbf{R$\to$R} & \textbf{$\to$G} & \textbf{$\to$S} & \textbf{$\to$R} & \textbf{$\to$G} & \textbf{$\to$S} & \textbf{$\to$R} \\
\midrule
gpt-4o & 93.9\,{\tiny \textcolor{gray01}{+2.9}} & \textcolor{blue}{90.7}\,{\tiny \textcolor{gray01}{+4.0}} & 92.0\,{\tiny \textcolor{gray01}{+5.1}} & \textcolor{blue}{95.1}\,{\tiny \textcolor{gray01}{+4.1}} & \textcolor{blue}{91.2}\,{\tiny \textcolor{gray01}{+4.5}} & 93.1\,{\tiny \textcolor{gray01}{+6.1}} & \textcolor{blue}{94.7}\,{\tiny \textcolor{gray01}{+3.7}} & \textcolor{blue}{90.4}\,{\tiny \textcolor{gray01}{+3.7}} & 92.5\,{\tiny \textcolor{gray01}{+5.6}} \\
gemini-2.5-flash & \textcolor{blue}{95.5}\,{\tiny \textcolor{gray01}{+1.6}} & 90.4\,{\tiny \textcolor{gray01}{+1.9}} & \textcolor{blue}{93.6}\,{\tiny \textcolor{gray01}{+1.3}} & 94.0\,{\tiny \textcolor{gray01}{+0.1}} & 90.0\,{\tiny \textcolor{gray01}{+1.5}} & \textcolor{blue}{93.2}\,{\tiny \textcolor{gray01}{+0.9}} & 94.4\,{\tiny \textcolor{gray01}{+0.5}} & 89.3\,{\tiny \textcolor{gray01}{+0.8}} & \textcolor{blue}{94.7}\,{\tiny \textcolor{gray01}{+2.4}} \\
claude-sonnet-4 & 91.7\,{\tiny \textcolor{gray01}{+0.3}} & 85.6\,{\tiny \textcolor{gray01}{+0.8}} & 90.7\,{\tiny \textcolor{gray01}{+4.8}} & 90.0\,{\tiny \textcolor{gray01}{$-$1.4}} & 86.8\,{\tiny \textcolor{gray01}{+2.0}} & 90.8\,{\tiny \textcolor{gray01}{+4.9}} & 90.4\,{\tiny \textcolor{gray01}{$-$1.0}} & 86.7\,{\tiny \textcolor{gray01}{+1.9}} & 90.1\,{\tiny \textcolor{gray01}{+4.3}} \\
grok-4-1-fast-reasoning & 91.5\,{\tiny \textcolor{gray01}{$-$0.5}} & 85.9\,{\tiny \textcolor{gray01}{+1.3}} & 89.1\,{\tiny \textcolor{gray01}{$-$1.1}} & 92.0\,{\tiny \textcolor{gray01}{\phantom{$-$}0.0}} & 85.7\,{\tiny \textcolor{gray01}{+1.2}} & 82.5\,{\tiny \textcolor{gray01}{$-$7.6}} & 93.1\,{\tiny \textcolor{gray01}{+1.1}} & 85.3\,{\tiny \textcolor{gray01}{+0.8}} & 82.4\,{\tiny \textcolor{gray01}{$-$7.7}} \\
\midrule
Qwen2-VL-7B-Instruct & 65.6\,{\tiny \textcolor{gray01}{$-$2.4}} & 57.9\,{\tiny \textcolor{gray01}{$-$0.5}} & 64.5\,{\tiny \textcolor{gray01}{$-$0.5}} & 64.0\,{\tiny \textcolor{gray01}{$-$4.0}} & 57.7\,{\tiny \textcolor{gray01}{$-$0.7}} & 64.0\,{\tiny \textcolor{gray01}{$-$1.1}} & 65.6\,{\tiny \textcolor{gray01}{$-$2.4}} & 57.6\,{\tiny \textcolor{gray01}{$-$0.8}} & 62.4\,{\tiny \textcolor{gray01}{$-$2.7}} \\
Llama-3.2-11B-Vision & 71.7\,{\tiny \textcolor{gray01}{+9.6}} & 64.3\,{\tiny \textcolor{gray01}{+18.4}} & 65.9\,{\tiny \textcolor{gray01}{+5.6}} & 73.1\,{\tiny \textcolor{gray01}{+10.9}} & 52.9\,{\tiny \textcolor{gray01}{+7.1}} & 61.9\,{\tiny \textcolor{gray01}{+1.6}} & 70.7\,{\tiny \textcolor{gray01}{+8.5}} & 56.5\,{\tiny \textcolor{gray01}{+10.7}} & 65.1\,{\tiny \textcolor{gray01}{+4.8}} \\
LLaVA-OneVision-1.5-8B-Instruct & 69.6\,{\tiny \textcolor{gray01}{$-$1.6}} & 64.8\,{\tiny \textcolor{gray01}{+1.3}} & 62.7\,{\tiny \textcolor{gray01}{+1.6}} & 70.8\,{\tiny \textcolor{gray01}{$-$0.4}} & 64.5\,{\tiny \textcolor{gray01}{+1.1}} & 62.9\,{\tiny \textcolor{gray01}{+1.9}} & 69.9\,{\tiny \textcolor{gray01}{$-$1.3}} & 65.1\,{\tiny \textcolor{gray01}{+1.6}} & 63.5\,{\tiny \textcolor{gray01}{+2.4}} \\
InternVL2\_5-26B & 45.9\,{\tiny \textcolor{gray01}{$-$13.1}} & \textcolor{red}{38.7}\,{\tiny \textcolor{gray01}{$-$6.9}} & \textcolor{red}{53.3}\,{\tiny \textcolor{gray01}{$-$6.4}} & \textcolor{red}{45.2}\,{\tiny \textcolor{gray01}{$-$13.7}} & \textcolor{red}{39.2}\,{\tiny \textcolor{gray01}{$-$6.4}} & 53.3\,{\tiny \textcolor{gray01}{$-$6.4}} & \textcolor{red}{45.6}\,{\tiny \textcolor{gray01}{$-$13.3}} & \textcolor{red}{37.1}\,{\tiny \textcolor{gray01}{$-$8.5}} & 52.3\,{\tiny \textcolor{gray01}{$-$7.5}} \\
Kimi-VL-A3B-Instruct & 47.5\,{\tiny \textcolor{gray01}{$-$3.7}} & 46.9\,{\tiny \textcolor{gray01}{$-$1.6}} & 59.5\,{\tiny \textcolor{gray01}{$-$0.3}} & 51.7\,{\tiny \textcolor{gray01}{+0.5}} & 48.7\,{\tiny \textcolor{gray01}{+0.1}} & \textcolor{red}{48.9}\,{\tiny \textcolor{gray01}{$-$10.8}} & 49.9\,{\tiny \textcolor{gray01}{$-$1.3}} & 49.1\,{\tiny \textcolor{gray01}{+0.5}} & \textcolor{red}{47.2}\,{\tiny \textcolor{gray01}{$-$12.5}} \\
Qwen2-VL-72B-Instruct & \textcolor{red}{33.6}\,{\tiny \textcolor{gray01}{$-$49.9}} & 70.9\,{\tiny \textcolor{gray01}{$-$8.5}} & 80.3\,{\tiny \textcolor{gray01}{\phantom{$-$}0.0}} & 51.7\,{\tiny \textcolor{gray01}{$-$31.8}} & 63.3\,{\tiny \textcolor{gray01}{$-$16.1}} & 75.9\,{\tiny \textcolor{gray01}{$-$4.4}} & 47.5\,{\tiny \textcolor{gray01}{$-$36.1}} & 70.1\,{\tiny \textcolor{gray01}{$-$9.3}} & 78.7\,{\tiny \textcolor{gray01}{$-$1.6}} \\
\bottomrule
\end{tabular}
\caption{Few-Shot Lift (FSL): accuracy with $k=3$ examples (subscripts show lift from zero-shot in \textcolor{gray01}{gray}). \textcolor{blue}{Blue} indicates highest accuracy, \textcolor{red}{red} indicates lowest accuracy per column.}

\label{tab:exp3_fsl}
\end{table*}

\begin{figure*}[t]
\centering
\includegraphics[width=\textwidth,height=0.4\textheight,keepaspectratio]{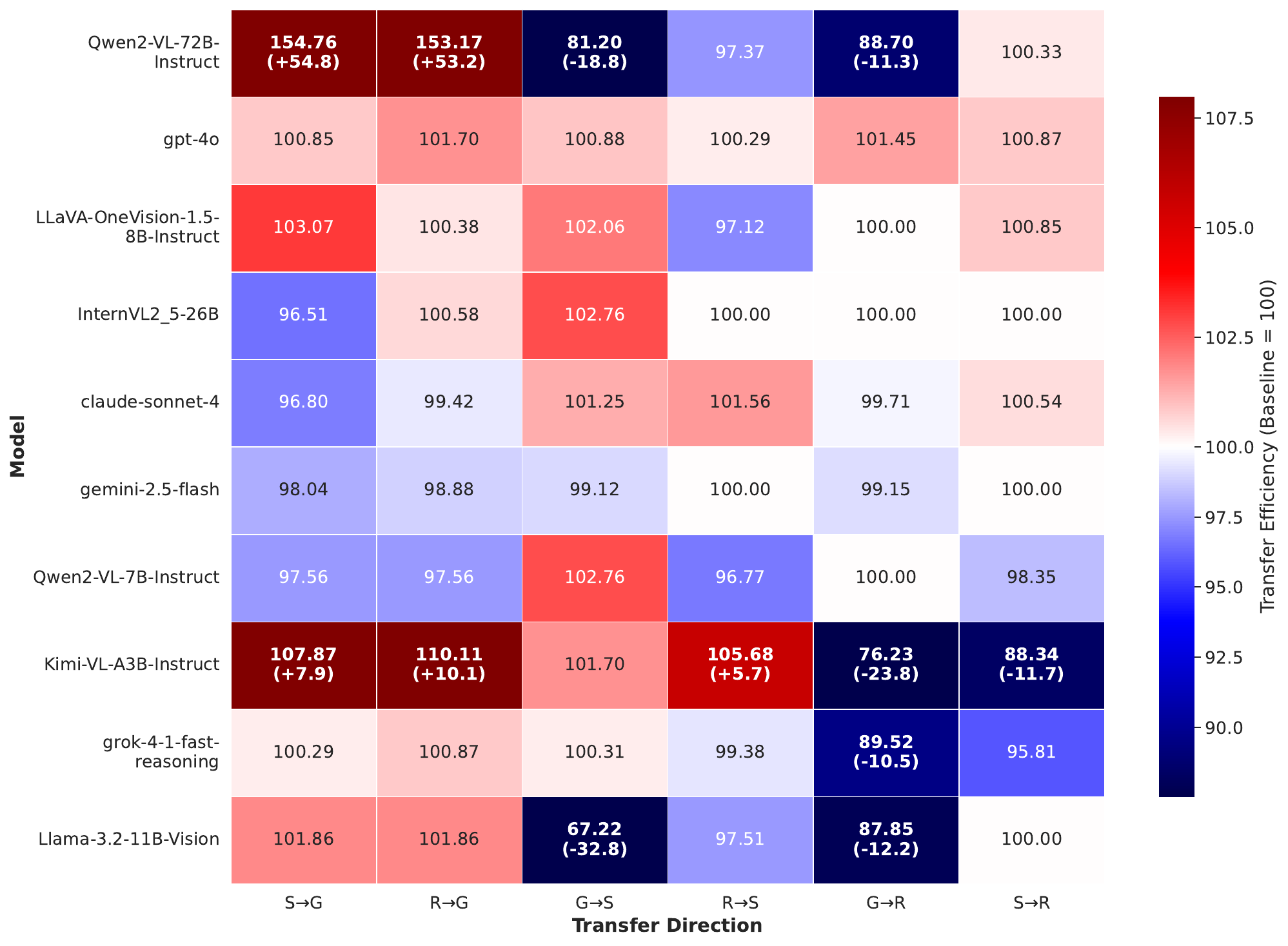}
\caption{Transfer Efficiency (TE) heatmap (baseline=100\%).}

\label{fig:exp3_te}
\end{figure*}

\section{Results and Analysis}

Our systematic evaluation of 10 state-of-the-art VLMs across the 375-instance evaluation split reveals a consistent and statistically significant finding: 
\textbf{for Punjabi, current multilingual systems are not truly multi-script}. Orthographic variation systematically fragments model performance despite constant semantic content and visual grounding, with gaps confirmed significant for 8 of 10 models.

\subsection{The Script Gap is Universal and Substantial}

\begin{figure}[H]
\centering
\includegraphics[width=1\columnwidth]{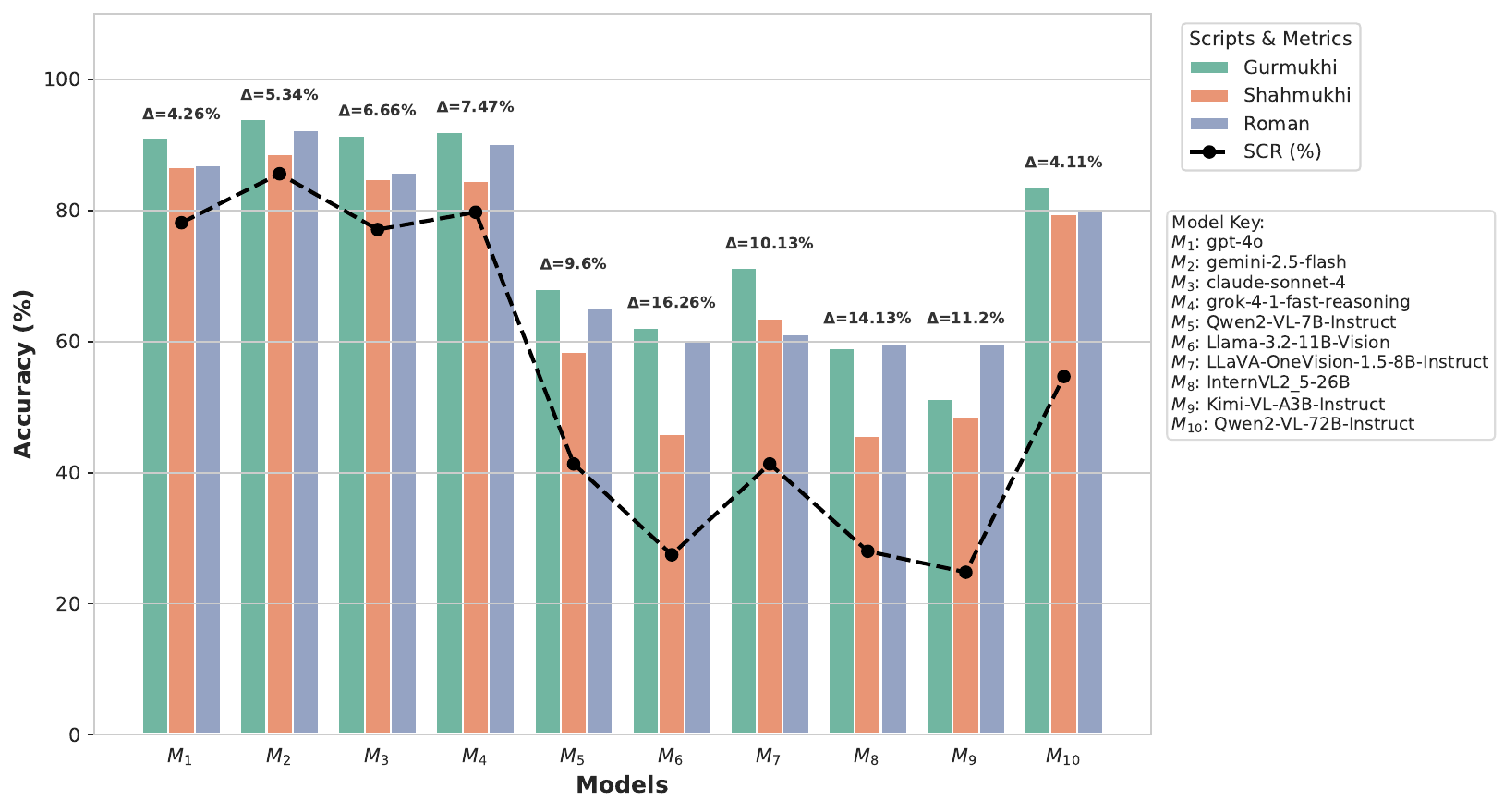}
\caption{Script-dependent accuracy and Script Consistency Rate (SCR) across 10 VLMs.}

\label{fig:exp1}
\end{figure}

Figure~\ref{fig:exp1} reveals our primary finding: \textbf{all evaluated models exhibit script-dependent performance degradation}, detailed in the per-model breakdown (\cref{app:results_tables}). Accuracy deltas range from 4.11\% (Qwen2-VL-72B-Instruct) to 16.26\% (Llama-3.2-11B-Vision). Critically, this affects frontier models, gpt-4o, gemini-2.5-flash, and claude-sonnet-4 demonstrate $\Delta$ values of 4.26\%, 5.34\%, and 6.66\% respectively, proving script bias is \textit{not} a low-resource artifact but a systematic limitation of current VLM designs.

The \textbf{Script Consistency Rate (SCR)} exposes the severity of orthographic fragmentation. gpt-4o achieves 90.93\% in Gurmukhi yet records only 78.13\% SCR, \textbf{12.8\% of its correctly answered instances are orthographically inconsistent}. This pattern intensifies in open-weights models: Llama-3.2-11B-Vision's SCR of 27.47\% means nearly three-quarters of instances cannot be solved consistently across scripts, making reported Punjabi accuracy figures unreliable for real-world multi-script users.

To verify that SCR captures genuine cross-script inconsistency rather than simply reflecting uniform per-item difficulty, we compare observed SCR against an independence baseline (the product of per-script accuracies, i.e. the SCR expected if script-wise correctness were statistically independent). Observed SCR exceeds this independence baseline for every model, by an average of 10.2 points (range: 1.4--15.5 points), indicating that correctness across scripts is positively correlated rather than independent, a model's success in one script is not orthogonal to its success in another on the same item. The gap is smallest for Qwen2-VL-72B-Instruct (1.4 points), one of the strongest open-weights models in raw accuracy (80--83\% across scripts); its high accuracy leaves less headroom for the independence-baseline gap to appear, consistent with a ceiling effect rather than weaker consistency.

Eight of ten models peak in Gurmukhi, reflecting its predominance in Indian web corpora. Shahmukhi consistently underperforms, averaging 7.7\% below Gurmukhi, indicating that Perso-Arabic cursive representations remain systematically undertrained even in frontier systems.

\begin{table}[t]
\centering
\footnotesize
\setlength{\tabcolsep}{2pt}
\renewcommand{\arraystretch}{1.2}
\begin{tabular}{@{}p{2.25 cm} cc cc cc@{}}
\toprule
\textbf{Model} & \multicolumn{2}{c}{\textbf{Gurmukhi}} & \multicolumn{2}{c}{\textbf{Shahmukhi}} & \multicolumn{2}{c}{\textbf{Roman}} \\
\cmidrule(lr){2-3} \cmidrule(lr){4-5} \cmidrule(lr){6-7}
& \textbf{T} & \textbf{VG} & \textbf{T} & \textbf{VG} & \textbf{T} & \textbf{VG} \\
\midrule
gpt-4o & 74.1 & 16.8 & \textcolor{blue}{69.9} & 16.8 & 62.4 & 24.5 \\
gemini-2.5-flash & 71.2 & 22.7 & 67.7 & 20.8 & 66.9 & 25.3 \\
claude-sonnet-4 & \textcolor{blue}{74.3} & 17.2 & 64.5 & 20.2 & 58.4 & 27.4 \\
grok-4-1-fast-reasoning & 73.6 & 18.4 & 63.2 & 21.3 & \textcolor{blue}{67.2} & \textcolor{red}{22.9} \\
\midrule
Qwen2-VL-7B-Instruct & 40.8 & 27.2 & 37.1 & 21.3 & 34.9 & \textcolor{blue}{30.1} \\
Llama-3.2-11B-Vision & 46.9 & 15.2 & 38.4 & \textcolor{red}{7.5} & 34.7 & 25.6 \\
LLaVA-OneVision-1.5-8B-Instruct & 48.8 & 22.4 & 45.3 & 18.1 & 37.1 & 24.0 \\
InternVL2\_5-26B & \textcolor{red}{36.8} & 22.1 & \textcolor{red}{32.3} & 13.3 & \textcolor{red}{32.8} & 26.9 \\
Kimi-VL-A3B-Instruct & 38.1 & \textcolor{red}{13.1} & 37.1 & 11.5 & 35.7 & 24.0 \\
Qwen2-VL-72B-Instruct & 47.5 & \textcolor{blue}{36.1} & 53.1 & \textcolor{blue}{26.4} & 51.7 & 28.5 \\
\bottomrule
\end{tabular}
\caption{Modality ablation showing Text-Only accuracy (T) and Visual Gain (VG). \textcolor{blue}{Blue} indicates highest, \textcolor{red}{red} indicates lowest values per column.}

\label{tab:exp2}
\end{table}

To confirm these gaps exceed chance variation, we applied McNemar's test to all pairwise script comparisons per model (Table~\ref{tab:mcnemar_full}, Appendix~\ref{app:mcnemar}). The Gurmukhi-Shahmukhi gap reaches statistical significance for 8 of 10 models, with 5 models at $p<0.001$. This holds across both frontier and open-weights systems, confirming that script bias is not a low-resource artifact. Qwen2-VL-72B-Instruct is the only model where no pairwise comparison reaches significance (minimum $p=0.131$), consistent with its anomalous few-shot profile discussed in Section~\ref{sec:exp3_results}. Non-significant comparisons - for example, GPT-4o Shah vs. Rom ($p=0.883$) and claude-sonnet-4 Shah vs. Rom ($p=0.892$) - are not overclaimed; they reflect genuine script-pair similarity for those models. Taken together, the McNemar results establish the Script Gap as statistically robust and not an artifact of dataset size.

\subsection{Visual Grounding Provides Parallel, Not Compensatory Gain}

Table~\ref{tab:exp2} reveals that visual information boosts performance uniformly (VG: 7.5--36.1\%) but \textbf{does not close the script gap}. gpt-4o's VG is nearly identical across Gurmukhi (16.8\%) and Shahmukhi (16.8\%), yet absolute accuracies differ by 4.3\%. Images provide \textbf{additive benefit}, not \textbf{compensatory repair}, the underlying orthographic fragmentation persists in multimodal settings.

Roman script shows the highest average VG (25.9\%) despite often being the best-performing script in zero-shot settings. This pattern suggests models lean on memorized surface patterns in high-resource scripts while requiring visual evidence when orthographic priors are weaker. Llama-3.2-11B-Vision's substantially low Shahmukhi VG (7.5\%) indicates \textbf{script-specific visual grounding failure}: even with images, it cannot integrate Perso-Arabic text effectively.

Notably, the Text-Only column of Table~\ref{tab:exp2} already exhibits a substantial script gap on its own, for gpt-4o, Text-Only accuracy differs by 11.7 points between Gurmukhi (74.1\%) and Roman (62.4\%), a larger gap than the corresponding multimodal delta. This indicates the Script Gap is not a purely multimodal phenomenon: it is already present in text-only comprehension and is carried into, rather than created by, the multimodal setting.

\subsection{Limited Cross-Script Transfer Suggests Script-Dependent Knowledge}
\label{sec:exp3_results}

Table~\ref{tab:exp3_fsl} and Figure~\ref{fig:exp3_te} reveal that in-context knowledge does not transfer reliably across scripts, indicating script-locked knowledge representation.

\paragraph{Negative Few-Shot Lift Exposes Brittleness.}
Several models exhibit \textit{performance degradation} when provided with in-context examples, revealing instability rather than adaptation under few-shot prompting. Most strikingly, Qwen2-VL-72B-Instruct shows a G$\to$G Few-Shot Lift (FSL) of \textbf{$-$49.9\%}: zero-shot accuracy drops from 83.5\% to 33.6\% when conditioned on three Gurmukhi exemplars. This catastrophic decline suggests severe in-context brittleness under low-resource script conditions. A plausible explanation is that Gurmukhi's comparatively limited representation in multimodal pretraining data prevents the model from developing stable in-context learning pathways for this orthography. Consequently, Gurmukhi exemplars may introduce tokenization-level ambiguity or activate conflicting attention patterns, causing contextual interference rather than task grounding. Importantly, the same model exhibits substantially more stable few-shot behavior in Shahmukhi and Roman Punjabi (FSL of $-$8.5\% and 0.0\%, respectively), indicating that the failure is script-specific rather than a general limitation of in-context learning.

\paragraph{Transfer Efficiency Reveals Anchor-Script Asymmetry.}
Transfer Efficiency (TE) further exposes asymmetries in cross-script 
in-context transfer behavior. Frontier models achieve near-perfect TE scores (98--101\%), suggesting robust script-agnostic transfer under in-context prompting. In contrast, Qwen2-VL-72B-Instruct exhibits highly asymmetric transfer dynamics: TE$_{\text{S}\to\text{G}}$=154.76\% and TE$_{\text{R}\to\text{G}}$=153.17\% indicate that cross-script exemplars outperform same-script prompting for Gurmukhi inference, while TE$_{\text{G}\to\text{S}}$=81.20\% reflects a substantial 19\% degradation in transfer efficiency in the reverse direction. Similar asymmetry appears in Llama-3.2-11B-Vision, where TE$_{\text{G}\to\text{S}}$=67.22\% corresponds to a \textbf{33\% reduction} in cross-script transfer efficiency. Together, these results are consistent with certain scripts acting as stronger 
``anchor scripts'' for in-context knowledge transfer, suggestive of script-specific 
internal representations rather than shared semantic ones. However, our experiments 
do not establish this causally: observed script-dependent gaps are also consistent 
with simpler confounds, including uneven pretraining data frequency across scripts, 
tokenizer coverage disparities, OCR difficulty, and general corpus imbalance (e.g., 
Gurmukhi's stronger representation in Indian web corpora relative to Shahmukhi, 
noted in Section~5.1). Disentangling script-locked representation from these 
confounds requires controlled studies on open-weights models with known pretraining 
mixtures, which we leave to future work.



\section{Conclusion}
We identify a critical blind spot in multilingual VLM evaluation: the ``One Language, One Script'' paradigm. Through PuMVR's 1000 parallel-script instances, we provide the first systematic evidence for Punjabi that state-of-the-art models exhibit substantial script-dependent bias: accuracy deltas reach 16\%, SCR falls to 24.8\%, and Transfer Efficiency drops below 67\%. Visual grounding provides additive benefit but does not close the script gap, confirming the bias is systematic and not a modality artifact.

We propose SCR as a mandatory metric for script-agnostic evaluation and provide a methodology transferable to dozens of multi-script languages. Achieving equitable AI coverage requires moving beyond language breadth to orthographic robustness.



\section*{Limitations}

\paragraph{Dataset Scope.} Our benchmark contains 1,000 curated instances across various reasoning dimensions. Experiments were conducted on the 375-instance evaluation split described in Section 4; all reported statistics are computed exclusively on this split. While this represents the first systematic cross-script multimodal evaluation, the size constrains fine-grained statistical analysis and represents a focused subset of reasoning capabilities. We prioritized quality and semantic equivalence over scale; future work should expand instance counts while maintaining our rigorous parallel-script methodology. Approximately 95\% of PuMVR images are AI-generated (Ethics Statement); while manually reviewed for coherence and accuracy, generation artifacts remain a possible confound for visual reasoning that a fully photographic benchmark would not carry.

\paragraph{Linguistic Generalizability.} We focus on Punjabi, whose three scripts provide typological diversity and minimal corpus overlap. However, script bias patterns may differ for: (1) languages with closer orthographic relationships (Serbian Cyrillic/Latin), (2) logographic systems (Chinese traditional/simplified), or (3) scripts with more balanced web representation. Our methodology is transferable, but empirical validation across diverse language families is needed before broad generalization. We provide a blueprint, not a universal law.

\paragraph{Evaluation Settings.} We assess zero-shot and few-shot performance without fine-tuning. Script bias may behave differently under full fine-tuning with script-balanced data, instruction tuning for cross-script robustness, or retrieval-augmented architectures. Our findings reflect contemporary deployment settings but may not predict behavior under targeted mitigation strategies.

\paragraph{Few-Shot Sensitivity.} We have not conducted a systematic sensitivity analysis of the Experiment~3 results to prompt wording or exemplar choice (Section~\ref{sec:exp3}). Some observed few-shot effects are large, including a 49.9-point accuracy drop for Qwen2-VL-72B-Instruct under monoscript Gurmukhi conditioning, and prompt or exemplar sensitivity is a plausible contributing factor. We verified these results reflect our implementation correctly, but present them as evidence of transfer behavior under our specific prompting setup rather than as prompt-independent estimates.


\paragraph{Model Coverage.} We evaluate 10 contemporary VLMs representing current state-of-the-art. Findings may not generalize to future architectures with improved cross-script mechanisms, proprietary systems with undisclosed script-balancing, or domain-specific models. Pretraining corpus composition is not publicly disclosed for the proprietary models evaluated here, so we cannot directly measure how much of the observed Script Gap is attributable to pretraining data frequency per script; controlled studies on open-weights models with known pretraining mixtures would help isolate this factor. Our work establishes a baseline for measuring progress rather than immutable limitations.

\paragraph{Cultural Entanglement.} Despite rigorous efforts at semantic equivalence, some instances may carry subtle script-specific cultural associations (Gurmukhi/Sikh, Shahmukhi/Islamic contexts) that are difficult to fully disentangle in a culturally grounded benchmark. This reflects the lived reality of script-culture associations but may introduce confounds beyond pure orthographic variation.

\section*{Ethics Statement}


\paragraph{Cultural Authenticity and Representation.} PuMVR was developed by native Punjabi speakers fluent in all three scripts and reviewed by community members. This ensured linguistic accuracy, cultural authenticity, and avoidance of stereotyping. Our team’s lived experience with script-dependent technological barriers informed the work but may limit perspective on other multi-script contexts.

\paragraph{Data Sources and Privacy.} Approximately 95\% of images in PuMVR are AI-generated (synthetic), used to ensure cultural specificity and clear licensing; the remainder are sourced from public domain archives, Wikimedia Commons, or original photography. AI-generated images were manually reviewed by curators for visual coherence and cultural accuracy before inclusion. No personally identifiable information is included. Human subjects in photographs are from public historical archives or are AI-generated.

\paragraph{Dual-Use Considerations.} This research exposes systemic bias to promote equitable AI access for multi-script communities. However, we acknowledge potential misuse: script-dependent performance insights could facilitate targeted disinformation or discrimination against low-resource script users. Our work diagnoses rather than optimizes such biases; we advocate for script-agnostic improvements that benefit all orthographies equally. We believe the benefits, improved fairness for over a billion users, substantially outweigh the risks.

\paragraph{Environmental Impact.} Total evaluation required approximately 250 Nvidia GH200 GPU hours, consuming an estimated 193 kWh of electricity and producing about 91.4 kg CO$_2$ (equivalent to 368 km of passenger vehicle travel).\footnote{Estimates assume a 700W average power draw for the GH200 Superchip and a PUE of 1.1.} While non-trivial, this cost is reported transparently and reflects current practices for large-scale multimodal evaluation, with potential benefits for improvingc equity across over one billion users of multi-script languages.

\paragraph{Transparency.} We have released all data, code, and detailed experimental protocols to enable reproducibility and community validation of our findings.

\section*{Acknowledgments}
We thank Japji Kaur for her contributions to the curation and annotation of PuMVR instances, and Fiza Asif Javed and Muhammad Mughees Ul Haq for their careful work as quality-check annotators. Their efforts were essential to the quality and reliability of this benchmark.

\newpage


\bibliography{custom}



\clearpage
\appendix
\crefalias{section}{appendix}

\section{Annotation Protocol Details}
\label{app:annotation_protocol}

Each instance was assessed across five dimensions: (i)~semantic equivalence across all three scripts, (ii)~answer correctness, verified by independently inspecting the image and question without consulting the provided answers, and script accuracy for (iii)~Gurmukhi, (iv)~Shahmukhi, and (v)~Roman Punjabi. Annotators were instructed to form independent judgments before recording any label and to apply a conservative criterion, marking \emph{No} in cases of genuine uncertainty rather than defaulting to \emph{Yes}. This conservative protocol is important for interpreting the results: the near-universal agreement observed reflects genuine expert consensus reached under an instruction regime explicitly designed to surface disagreement, rather than annotator passivity or a low-effort default. A stratified sample of 375~instances drawn from the full benchmark was selected to match the evaluation split used in all three experiments, enabling agreement figures to be interpreted directly with respect to experimental coverage.

\section{Model Configurations}

We provide detailed specifications for all evaluated models to ensure reproducibility. Table~\ref{tab:model-config} summarizes the key hyperparameters used across all experiments.

\begin{table*}[t]
\centering
\renewcommand{\arraystretch}{1.25}
\small
\setlength{\tabcolsep}{4pt}
\begin{tabular}{>{\raggedright\arraybackslash}p{3.5cm} r c c c c c l}
\toprule
\textbf{Model} & \textbf{Organization} & \textbf{Params} & \textbf{Precision} & \textbf{Temp.} & \textbf{Max Tok.} & \textbf{Seed} & \textbf{Inference} \\
\midrule
\multicolumn{8}{l}{\textcolor{blue01}{{\textit{Proprietary Frontier Models}}}} \\
\midrule
gpt-4o & OpenAI & -- & Native & 0.1 & 128 & 42 & API \\
gemini-2.5-flash & Google DeepMind & -- & Native & 0.1 & 128 & 42 & Vertex AI \\
claude-sonnet-4 & Anthropic & -- & Native & 0.1 & 128 & -- & Vertex AI \\
grok-4-1-fast-reasoning & xAI & -- & Native & 0.1 & 128 & 42 & API \\
\midrule
\multicolumn{8}{l}{\textcolor{blue01}{{\textit{Open-Weights Models}}}} \\
\midrule
Qwen2-VL-7B-Instruct & Alibaba & 7B & bfloat16 & 0.1 & 128 & 42 & Local \\
Qwen2-VL-72B-Instruct & Alibaba & 72B & bfloat16 & 0.1 & 128 & 42 & Local \\
Llama-3.2-11B-Vision & Meta & 11B & bfloat16 & 0.1 & 128 & 42 & Local \\
LLaVA-OneVision-1.5-8B-Instruct & LMMS-Lab & 8B & bfloat16 & 0.1 & 128 & 42 & Local \\
InternVL2\_5-26B & OpenGVLab & 26B & bfloat16 & 0.1 & 128 & 42 & Local \\
Kimi-VL-A3B-Instruct & Moonshot AI & 3B & auto & 0.2 & 128 & 42 & Local \\
\bottomrule
\end{tabular}
\caption{Model configuration summary. All models use greedy decoding (temperature $\leq$ 0.2). ``Local'' denotes inference on NVIDIA GH200 with CUDA.}
\label{tab:model-config}
\end{table*}

\subsection{Hardware Infrastructure}

\noindent\textbf{API-Based Models:} All proprietary models were accessed via their respective APIs with no local computational requirements. claude-sonnet-4 was accessed through Google Cloud's Vertex AI service (us-east5 region).

\noindent\textbf{Local Models :} Evaluated on NVIDIA GH200 Grace Hopper Superchip with:
\begin{itemize}
    \item \textbf{GPU Memory:} 120GB HBM3
    \item \textbf{CUDA Version:} 12.1+
    \item \textbf{Framework:} PyTorch 2.1.0+
    \item \textbf{Memory Management:} \texttt{device\_map = "cuda" } with \texttt{max\_memory = \{0: "110GB"\}}
    \item \textbf{Optimization:} \texttt{low\_cpu\_mem\_usage = True}, \texttt{trust\_remote\_code = True}
\end{itemize}

\subsection{Model-Specific Implementation Details}

\noindent\textbf{Qwen2-VL Series (7B \& 72B):}
\begin{itemize}
    \item \textbf{Architecture:} \texttt{Qwen2VLForConditionalGeneration}
    \item \textbf{Tokenizer:} \texttt{AutoProcessor} with custom Qwen2 tokenizer
    \item \textbf{Image Processing:} Automatic via processor's \texttt{apply\_chat\_template}
    \item \textbf{Input Format:} List-of-dicts conversation structure with image and text content blocks
    \item \textbf{Generation:} Output token trimming (remove input length from generated sequences)
\end{itemize}

\noindent\textbf{Llama-3.2-11B-Vision:}
\begin{itemize}
    \item \textbf{Architecture:} \texttt{MllamaForConditionalGeneration}
    \item \textbf{Processor:} \texttt{AutoProcessor} (handles \texttt{MllamaProcessor} automatically)
    \item \textbf{Image Handling:} Passed to processor via \texttt{images} parameter
    \item \textbf{Special Tokens:} \texttt{<|image|>} tokens inserted automatically by \texttt{apply\_chat\_template}
    \item \textbf{Decoding:} Input tokens trimmed from output: \texttt{output\_ids[len(input\_ids):]}
\end{itemize}

\noindent\textbf{LLaVA-OneVision-1.5-8B-Instruct:}
\begin{itemize}
    \item \textbf{Architecture:} \texttt{AutoModelForCausalLM} (requires \texttt{trust\_remote\_code=True})
    \item \textbf{Vision Backbone:} Custom \texttt{rice\_vit} architecture
    \item \textbf{Processor:} \texttt{AutoProcessor} with batch processing
    \item \textbf{Input Format:} Text list + image list with automatic alignment
    \item \textbf{Decoding:} \texttt{skip\_special\_tokens=True}, \texttt{clean\_up\_tokenization\_spaces=True}
\end{itemize}

\noindent\textbf{InternVL2\_5-26B:}
\begin{itemize}
    \item \textbf{Architecture:} \texttt{AutoModel} (requires \texttt{trust\_remote\_code=True})
    \item \textbf{Dynamic Resolution:} Custom preprocessing with aspect ratio preservation
    \item \textbf{Image Preprocessing:} BICUBIC interpolation to 448$\times$448 tiles (min\_num=1, max\_num=12)
    \item \textbf{Thumbnail Strategy:} Appended for multi-tile images (use\_thumbnail=True)
    \item \textbf{Generation:} Native \texttt{.chat()} method with pixel\_values tensor stack
    \item \textbf{Flash Attention:} Enabled via \texttt{use\_flash\_attn=True}
\end{itemize}

\noindent\textbf{Kimi-VL-A3B-Instruct:}
\begin{itemize}
    \item \textbf{Architecture:} \texttt{AutoModelForCausalLM}
    \item \textbf{Processor:} Dual tokenizer setup (main tokenizer + processor tokenizer for chat template)
    \item \textbf{Chat Template:} Applied via \texttt{processor.tokenizer.apply\_chat\_\\template()}
    \item \textbf{Input Structure:} Type-annotated content list: \texttt{\{"type": "image"\}}, \texttt{\{"type": "text"\}}
    \item \textbf{Temperature:} 0.2
    \item \textbf{Token Management:} \texttt{pad\_token\_id} fallback to \texttt{eos\_token\_id} if unavailable
\end{itemize}

\noindent\textbf{API Models (gpt-4o, gemini-2.5-flash, claude-sonnet-4, grok-4-1-fast-reasoning):}
\begin{itemize}
    \item \textbf{Image Encoding:} Base64 JPEG encoding (RGB conversion applied if input is RGBA/other modes)
    \item \textbf{gpt-4o:} OpenAI SDK with \texttt{data : image/jpeg;base64} URL format
    \item \textbf{gemini-2.5-flash:} Google Generative AI SDK with \texttt{[prompt\_text, pil\_image]} input list
    \item \textbf{Gemini Safety:} All harm categories set to \texttt{BLOCK\_NONE} to prevent refusal on research data
    \item \textbf{claude-sonnet-4:} Anthropic Vertex AI client with base64 source in content blocks
    \item \textbf{grok-4-1-fast-reasoning:} OpenAI-compatible SDK (\texttt{base\_url="https://api.x.ai/v1"}) with image\_url format
\end{itemize}

\subsection{Reproducibility Safeguards}

\noindent\textbf{Random Seeds:} All local models initialize with \texttt{seed=42}:
\begin{verbatim}
torch.manual_seed(42)
torch.cuda.manual_seed_all(42)
\end{verbatim}

\noindent\textbf{API Seeds:} Where supported (gpt-4o, gemini-2.5-flash, grok-4-1-fast-reasoning), \texttt{seed=42} parameter passed to API calls.

\noindent\textbf{Incremental Saving:} Results saved immediately after each prediction via CSV append mode to prevent data loss on system crashes.

\noindent\textbf{Resume Capability:} All evaluators check for existing CSV files and skip completed instance IDs, enabling seamless resumption after interruptions.

\noindent\textbf{Deterministic Decoding:} Temperature $\leq$ 0.2 with \texttt{do\_sample=(temperature > 0)} ensures greedy/near-greedy decoding for reproducibility.


\section{Complete Results Tables}
\label{app:results_tables}

\subsection{Experiment 1: Per-Script Accuracy Details}
\label{app:exp1_results}

Table~\ref{tab:exp1_detailed} presents the complete accuracy results across all three scripts for each model.


\begin{table}[ht]
\centering
\renewcommand{\arraystretch}{1.25}
\footnotesize
\setlength{\tabcolsep}{3pt} 
\begin{tabularx}{\columnwidth}{>{\RaggedRight\arraybackslash}X cccc} 
\toprule
\textbf{Model} & \textbf{Gur.} & \textbf{Shah.} & \textbf{Rom.} & \textbf{SCR} \\
\midrule
gpt-4o & 90.93 & 86.67 & 86.93 & 78.13 \\
gemini-2.5-flash & \textcolor{blue}{93.87} & \textcolor{blue}{88.53} & \textcolor{blue}{92.27} & \textcolor{blue}{85.60} \\
claude-sonnet-4 & 91.42 & 84.76 & 85.83 & 77.09 \\
grok-4-1-fast-reasoning & 92.00 & 84.53 & 90.13 & 79.73 \\
Qwen2-VL-7B-Instruct & 68.00 & 58.40 & 65.07 & 41.33 \\
Llama-3.2-11B-Vision & 62.13 & 45.87 & 60.27 & \textcolor{red}{27.47} \\
LLaVA-OneVision-1.5-8B-Instruct & 71.20 & 63.47 & 61.07 & 41.33 \\
InternVL2\_5-26B & 58.93 & \textcolor{red}{45.60} & \textcolor{red}{59.73} & 28.00 \\
Kimi-VL-A3B-Instruct & \textcolor{red}{51.20} & 48.53 & \textcolor{red}{59.73} & 24.80 \\
Qwen2-VL-72B-Instruct & 83.54 & 79.43 & 80.25 & 54.67 \\
\midrule
\textbf{Average} & \textbf{76.32} & \textbf{68.58} & \textbf{74.13} & \textbf{53.82} \\
\bottomrule
\end{tabularx}
\caption{Zero-shot accuracy (\%) across scripts. SCR = Script Consistency Rate.}
\label{tab:exp1_detailed}
\end{table}
\paragraph{Confidence Intervals.} Wilson score 95\% confidence intervals on per-script accuracy range from $\pm$2.5--3.7 points for frontier models to $\pm$3.8--5.0 points for open-weights models. As these are computed per script on the same 375 items rather than independent samples, overlap between them is not informative for paired significance; McNemar's test (Appendix~\ref{app:mcnemar}) is the appropriate test here and is unaffected by this.

\subsection{Experiment 3: Transfer Efficiency Matrix}
\label{app:exp3_results}

Table~\ref{tab:exp3_te} shows Transfer Efficiency (TE) values for all few-shot transfer conditions.



\section{Full Experimental Prompts}
\label{appendix:prompts}

This section provides the exact prompt templates used across all experiments. To ensure high-fidelity evaluation, we enforced strict formatting constraints.

\subsection{Notation and Placeholder Definitions}
To maintain clarity across the templates provided in this appendix, we use the following placeholders to denote dynamically injected content:

\begin{itemize}
    \item \texttt{\{question\}}: The Punjabi query translated into the specific target script (Gurmukhi, Shahmukhi, or Roman).
    \item \texttt{\{formatted\_options\}}: A list of four multiple-choice candidates. Each candidate is presented on a new line without alphabetical prefixes (A, B, C, D) or numerical markers.
    \item \texttt{\{script\_name\}}: The English name of the target script (e.g., ``Gurmukhi'', ``Shahmukhi'', or ``Roman'').
    \item \texttt{\{Script Instruction\}}: A script-specific directive used in Experiments to trigger internal script-specific reasoning. The values used were:
    \begin{itemize}
        \item \textbf{Gurmukhi:} ``{\gurmukhi ਗੁਰਮੁਖੀ ਵਿੱਚ}''
        \item \textbf{Shahmukhi:} ``{\shahmukhi شاہ مکھی وچ}''
        \item \textbf{Roman:} ``in Roman script''
    \end{itemize}
\end{itemize}

\subsection{Design Rationale: Text-Based vs. Letter-Based Selection}
In all experiments, models were instructed to output the \textit{exact text} of the correct option rather than a single identifier (e.g., "A", "B"). This design choice was made for two primary reasons:
\begin{enumerate}
    \item \textbf{Script Comprehension Verification:} A model might correctly guess a letter (25\% probability) without truly processing the script. Forcing the model to reproduce the script-specific text ensures it can parse and generate the target orthography.
    \item \textbf{Failure Mode Analysis:} By requiring text output, we could identify cases where models produced "hallucinated" characters or mixed scripts, data that would be lost if restricted to single-letter outputs.
\end{enumerate}

\subsection{Experiment 1: Baseline Script Gap}
The primary benchmark used the following template with English instructions and script-specific constraints.

\begin{tcolorbox}[colback=blue!5!white,colframe=blue!75!black,title=Exp 1: Standard Prompt,fonttitle=\bfseries]
\fontsize{10pt}{12pt}\selectfont
\texttt{Question: \{question\}}\\
\\
\texttt{Options:} \texttt{\{formatted\_options\}}\\
\\
\texttt{Task: Identify the correct option based on the image.}\\
\\
\texttt{Constraint: \{Answer in [Script Name] script.\} Copy the exact text of the correct option. Do not explain.}\\
\\
\texttt{CRITICAL RULES:}\\
\texttt{1. Answer MUST be in the SAME script as the question}\\
\texttt{2. Copy EXACTLY one option from above - character by character}\\
\texttt{3. NO explanations, NO extra words, NO English translations}\\
\texttt{4. NO letters like A), B), C) or numbers}\\
\texttt{5. Output ONLY the option text, nothing else}\\
\\
\texttt{Your answer (copy exact text from options):}
\end{tcolorbox}

\subsection{Experiment 2: Native Instruction Prompting}
This experiment used similar prompts as experiment 1, without the image input to test model performance and establish a baseline. 

\begin{tcolorbox}[colback=blue!5!white,colframe=blue!75!black,title=Exp 2: Native Script Instructions,fonttitle=\bfseries]
\fontsize{10pt}{12pt}\selectfont
\texttt{Question \{Script Instruction\}: \{question\}}\\
\\
\texttt{Options:} \texttt{\{formatted\_options\}}\\
\\
\texttt{CRITICAL RULES:}\\
\\
\texttt{1. Answer MUST be in the SAME script as the question}\\
\texttt{2. Copy EXACTLY one option from above - character by character}\\
\texttt{3. NO explanations, NO extra words, NO English translations}\\
\texttt{4. NO letters like A), B), C) or numbers}\\
\texttt{5. Output ONLY the option text, nothing else}\\
\\
\texttt{Your answer (copy exact text from options):}\\
\end{tcolorbox}

\subsection{Experiment 3: System Prompting (Few-Shot)}
For experiments involving system-level instructions, the following persona-based prompt was utilized.

\begin{tcolorbox}[colback=blue!5!white,colframe=blue!75!black,title=Exp 3: System Prompt,fonttitle=\bfseries, after skip=5mm]
\fontsize{10pt}{12pt}\selectfont
\texttt{You are a precise answering assistant. \\
You will be given a visual question and options. \\
You must output ONLY the exact text of the correct option in \{Script Name\} script.} \\
\\
\texttt{CRITICAL RULES:\\
1. NO reasoning.\\
2. NO explanations.\\
3. NO option letters (like A, B).\\
4. Output strictly the option text.}
\end{tcolorbox}

\section{Error Classification Details}
\label{app:error_analysis}

Table~\ref{tab:error_analysis} reports the per-script breakdown of error types across all 11,250 model responses (10 models $\times$ 375 instances $\times$ 3 scripts), confirming that the generative evaluation design measures comprehension rather than output formatting difficulty.

\begin{table}[H]
\centering
\small
\setlength{\tabcolsep}{3pt}
\renewcommand{\arraystretch}{1.1}
\begin{tabular}{lccc}
\toprule
\textbf{Script} & \textbf{Comp. Failures} & \textbf{Empty} & \textbf{Formatting} \\
\midrule
Gurmukhi  & 99.8\% & 0.2\% & 0.0\% \\
Shahmukhi & 99.3\% & 0.7\% & 0.0\% \\
Roman     & 99.7\% & 0.3\% & 0.0\% \\
\midrule
\textbf{Aggregate} & \textbf{99.59\%} & \textbf{0.41\%} & \textbf{0.00\%} \\
\bottomrule
\end{tabular}
\caption{Error classification across all model responses by script. Comprehension failures are complete wrong-option selections; formatting artifacts include mixed-script or hallucinated characters.}
\label{tab:error_analysis}
\end{table}

At the model level, claude-sonnet-4 showed the lowest comprehension failure rate at 96.55\%, with 3.45\% empty responses constituting the sole notable outlier. All other models showed comprehension failure rates at or above 98.77\%.

\section{McNemar Test Results}
\label{app:mcnemar}

Table~\ref{tab:mcnemar_full} reports McNemar's test results for all pairwise script comparisons across all 10 models. We apply McNemar's test treating per-instance correctness as paired binary observations across 375 instances. For each model, we test three script pairs: Gurmukhi vs. Shahmukhi, Gurmukhi vs. Roman, and Shahmukhi vs. Roman. The chi-squared variant is used when discordant pairs exceed 25; the exact variant otherwise.

Several patterns emerge. First, the Gurmukhi--Shahmukhi gap is the most consistently significant pair across models, confirming Shahmukhi as the most undertrained script. Second, the Gurmukhi--Roman gap is significant for some models (claude-sonnet-4, LLaVA) but not others, suggesting variable Roman script coverage across model families. Third, Qwen2-VL-72B-Instruct is the only model where no pair reaches significance, consistent with its more balanced cross-script performance observed in Table~\ref{tab:exp1_detailed}. These results validate that the Script Gap reported in Section~5.1 is statistically robust and not an artifact of the dataset size.

\begin{table}[h!]
\centering
\footnotesize
\setlength{\tabcolsep}{2.5pt}
\renewcommand{\arraystretch}{1.25}
\begin{tabular}{@{}p{3cm}ccc@{}}
\toprule
\textbf{Model} & \textbf{Script Pair} & \textbf{$p$-value} & \textbf{Sig.} \\
\midrule
gpt-4o & Gur vs Shah & 0.025 & * \\
gpt-4o & Gur vs Rom & 0.041 & * \\
gpt-4o & Shah vs Rom & 0.883 & ns \\
gemini-2.5-flash & Gur vs Shah & 0.001 & ** \\
gemini-2.5-flash & Gur vs Rom & 0.146 & ns \\
gemini-2.5-flash & Shah vs Rom & 0.038 & * \\
claude-sonnet-4 & Gur vs Shah & $<$0.001 & *** \\
claude-sonnet-4 & Gur vs Rom & 0.003 & ** \\
claude-sonnet-4 & Shah vs Rom & 0.892 & ns \\
grok-4-1-fast-reasoning & Gur vs Shah & $<$0.001 & *** \\
grok-4-1-fast-reasoning & Gur vs Rom & 0.265 & ns \\
grok-4-1-fast-reasoning & Shah vs Rom & 0.006 & ** \\
\midrule
Qwen2-VL-7B-Instruct & Gur vs Shah & $<$0.001 & *** \\
Qwen2-VL-7B-Instruct & Gur vs Rom & 0.343 & ns \\
Qwen2-VL-7B-Instruct & Shah vs Rom & 0.044 & * \\
Llama-3.2-11B-Vision & Gur vs Shah & $<$0.001 & *** \\
Llama-3.2-11B-Vision & Gur vs Rom & 0.608 & ns \\
Llama-3.2-11B-Vision & Shah vs Rom & $<$0.001 & *** \\
LLaVA-OneVision-1.5-8B-Instruct & Gur vs Shah & 0.004 & ** \\
LLaVA-OneVision-1.5-8B-Instruct & Gur vs Rom & $<$0.001 & *** \\
LLaVA-OneVision-1.5-8B-Instruct & Shah vs Rom & 0.415 & ns \\
InternVL2\_5-26B & Gur vs Shah & $<$0.001 & *** \\
InternVL2\_5-26B & Gur vs Rom & 0.857 & ns \\
InternVL2\_5-26B & Shah vs Rom & $<$0.001 & *** \\
Kimi-VL-A3B-Instruct & Gur vs Shah & 0.430 & ns \\
Kimi-VL-A3B-Instruct & Gur vs Rom & 0.006 & ** \\
Kimi-VL-A3B-Instruct & Shah vs Rom & 0.001 & ** \\
Qwen2-VL-72B-Instruct & Gur vs Shah & 0.131 & ns \\
Qwen2-VL-72B-Instruct & Gur vs Rom & 0.208 & ns \\
Qwen2-VL-72B-Instruct & Shah vs Rom & 0.892 & ns \\
\bottomrule
\end{tabular}
\caption{McNemar's test across all script pairs and models. *** $p{<}0.001$, ** $p{<}0.01$, * $p{<}0.05$, ns = not significant.}
\label{tab:mcnemar_full}
\end{table}

\begin{table*}[t]
\centering
\renewcommand{\arraystretch}{1.25}
\fontsize{10pt}{12pt}\selectfont
\setlength{\tabcolsep}{8pt}
\begin{tabular}{p{3.5cm} cccccc c}
\toprule
\textbf{Model} & \textbf{S$\rightarrow$G} & \textbf{R$\rightarrow$G} & \textbf{G$\rightarrow$S} & \textbf{R$\rightarrow$S} & \textbf{G$\rightarrow$R} & \textbf{S$\rightarrow$R} & \textbf{Avg. TE} \\
\midrule
gpt-4o & 100.85 & 101.70 & 100.88 & 100.29 & \textcolor{blue}{101.45} & \textcolor{blue}{100.87} & \textbf{101.01} \\
gemini-2.5-flash & 98.04 & 98.88 & 99.12 & 100.00 & 99.15 & 100.00 & 99.20 \\
claude-sonnet-4 & 96.80 & 99.42 & 101.25 & 101.56 & 99.71 & 100.54 & 99.88 \\
grok-4-1-fast-reasoning & 100.29 & 100.87 & 100.31 & 99.38 & 89.52 & 95.81 & 97.70 \\
\midrule
Qwen2-VL-7B-Instruct & 97.56 & \textcolor{red}{97.56} & \textcolor{blue}{102.76} & \textcolor{red}{96.77} & 100.00 & 98.35 & 98.83 \\
Llama-3.2-11B-Vision & 101.86 & 101.86 & \textcolor{red}{67.22} & 97.51 & 87.85 & 100.00 & \textcolor{red}{92.72} \\
LLaVA-OneVision-1.5-8B-Instruct & 103.07 & 100.38 & 102.06 & 97.12 & 100.00 & 100.85 & 100.58 \\
InternVL2\_5-26B & \textcolor{red}{96.51} & 100.58 & \textcolor{blue}{102.76} & 100.00 & 100.00 & 100.00 & 99.98 \\
Kimi-VL-A3B-Instruct & 107.87 & 110.11 & 101.70 & \textcolor{blue}{105.68} & \textcolor{red}{76.23} & \textcolor{red}{88.34} & 98.32 \\
Qwen2-VL-72B-Instruct & \textcolor{blue}{154.76} & \textcolor{blue}{153.17} & 81.20 & 97.37 & 88.70 & 100.33 & \textcolor{blue}{112.59} \\
\midrule
\textbf{Average} & \textbf{105.76} & \textbf{106.45} & \textbf{95.93} & \textbf{99.57} & \textbf{94.26} & \textbf{98.51} & \textbf{100.08} \\
\bottomrule
\end{tabular}
\caption{Transfer Efficiency (\%) across script pairs. Notation: G=Gurmukhi, S=Shahmukhi, R=Roman.}
\label{tab:exp3_te}
\end{table*}

\end{document}